\documentclass[letterpaper,10pt,conference]{ieeeconf}
\IEEEoverridecommandlockouts
\usepackage{amsmath,amssymb}
\usepackage{booktabs}
\usepackage{graphicx}
\usepackage{microtype}
\usepackage{flushend}
\usepackage{marvosym}

\renewcommand{\footnoterule}{%
  \kern-3pt
  \hrule width 0.4\columnwidth
  \kern 2.6pt
}

\title{\LARGE \bf Rethinking Learned Occupancy in Autonomous Active Mapping \\ with Observation-Gated Filtering}

\author{Jiahui Zhang$^{1}$, Bonian Han$^{1}$, Gongbo Liang$^{2}$, and Yu Zhang$^{1 \text{\Letter}}$%
\thanks{This work is supported by NSF Grant No. 2429968.}%
\thanks{$^{1}$Boise State University}%
\thanks{$^{2}$Texas A\&M University--San Antonio}%
\thanks{$^{\text{\Letter}}$Correspondence to: Yu Zhang: \texttt{yzhang@boisestate.edu}}}

\begin{document}
\maketitle
\thispagestyle{empty}
\pagestyle{empty}

\begin{abstract}
Autonomous 3D active mapping requires a space robot to choose where to sense while building the geometry needed for navigation. Learned occupancy completion extends spatial context beyond the current field of view, but one predicted map often serves two planning roles: it scores expected surface gain and constrains collision-free motion. Unsupported occupancy can therefore distort both where the robot looks and where it believes it can travel. We study this coupled interface in a controlled closed-loop benchmark by holding the active-mapping system fixed and varying only its planner-facing occupancy across observation-only, learned, oracle-corrected, and ground-truth conditions. Improving occupancy accuracy does not monotonically improve closed-loop coverage: across 25 starts, planning with ground-truth occupancy reaches 70\% of the learned baseline's final coverage 12.7 steps earlier on average, while increasing final coverage by only 0.031. Guided by this diagnosis, we introduce an observation-gated filter that retains completion in insufficiently observed regions and suppresses predictions only after repeated frustum exposure without nearby RGB-D support. The filter improves both targeted failure-prone starts without retraining or ground truth. These results motivate online revision of planner-facing geometry during autonomous intervals between communication windows. The current study assumes benchmark RGB-D observations and sufficiently accurate pose estimates; planetary sensing conditions and accumulated localization drift remain to be evaluated.
\end{abstract}

\section{Introduction}

Autonomous 3D active mapping couples reconstruction with action selection: at each step, a robot must update an incomplete map and choose the observation that will improve it next. This closed loop is especially consequential for planetary surface or subsurface reconnaissance, where supervision is delayed, motion is costly, and reliable external localization may be unavailable ~\cite{azkarate2020gnc}~\cite{sasaki2020where}. Learned geometric completion is attractive in this setting. An occupancy model can use partial observations to predict unobserved structure and guide sensing before a complete map exists~\cite{mescheder2019occupancy}~\cite{ramakrishnan2020occupancy}.

The same prior creates a coupled risk. In recent active-mapping systems, one occupancy map contributes to the expected new surface visible from a candidate viewpoint and also constrains collision-free motion ~\cite{guedon2023macarons}~\cite{li2026magician}. A hallucinated surface may therefore attract the robot as an apparently informative target, block a route through free space, or do both. When the mapper must operate between communication windows, planner-facing geometry must remain revisable as onboard evidence accumulates.

\begin{figure}[t]
\centering
\includegraphics[width=\columnwidth]{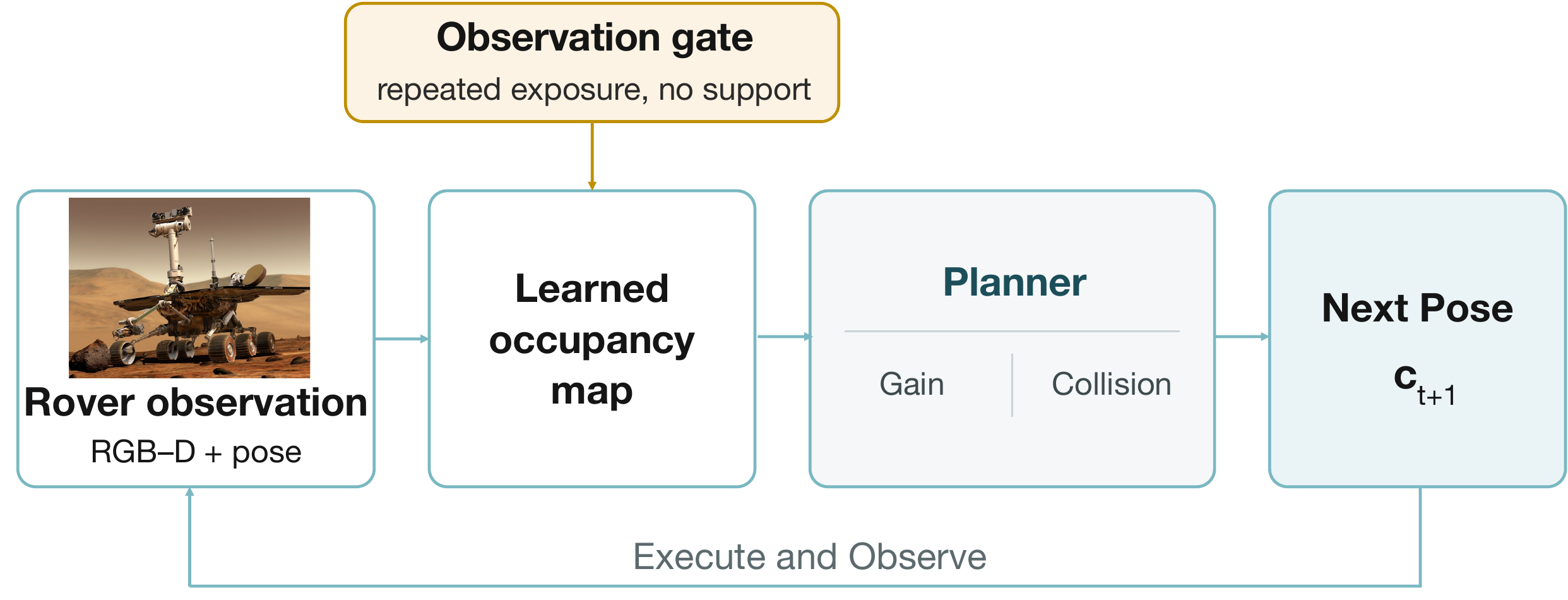}
\caption{\textbf{Closed-loop occupancy-guided active mapping.} Each rover observation updates a learned occupancy map used for both gain estimation and collision checking. Repeated exposure without nearby RGB-D support triggers observation-gated filtering before planning. The selected motion produces the next observation.}
\label{fig:loop}
\end{figure}

We therefore ask: \emph{when should an active mapper retain a learned geometric prior, and when should accumulated observations override it?} We use a controlled terrestrial 3D benchmark to isolate this planner-facing mechanism. The experiments do not claim validation under planetary appearance, degraded sensing, or long-horizon localization drift. They instead diagnose the shared occupancy interface and test an online revision rule for the closed perception, mapping, and planning loop in Fig.~\ref{fig:loop}.

\section{Related Work}

Space-robot autonomy couples onboard state estimation, guidance, hazard avoidance, and local motion planning so that a vehicle can continue operating when external localization or frequent ground intervention is unavailable~\cite{azkarate2020gnc}. Exploration determines where sensing effort should be spent. Coordinated rover--copter planning treats mapping location as a mission decision rather than a passive by-product of navigation~\cite{sasaki2020where}. Predictive perception has been studied for planetary exploration~\cite{otsu2017look}, while large-scale mapping addresses perceptually degraded subterranean environments~\cite{ebadi2020lamp}. Receding-horizon exploration and uncertainty-guided reconstruction similarly couple view utility with reachable motion~\cite{bircher2016nbv}~\cite{lee2022uncertainty}. These systems establish the importance of onboard spatial reasoning, but do not isolate failures caused when learned completion is shared by gain estimation and collision checking.

Learned occupancy provides a complementary capability. Occupancy Networks represent 3D geometry as a continuous inside--outside function ~\cite{mescheder2019occupancy}, while occupancy anticipation predicts free and occupied space for navigation before the environment is fully observed ~\cite{ramakrishnan2020occupancy}. For active reconstruction, SCONE and MACARONS use learned occupancy and coverage anticipation ~\cite{guedon2022scone}~\cite{guedon2023macarons}; NARUTO plans from reconstruction uncertainty~\cite{feng2024naruto}; and NextBestPath searches beyond a single view~\cite{li2025nextbestpath}. MAGICIAN converts pretrained occupancy into imagined Gaussian primitives for long-horizon planning ~\cite{li2026magician}. These approaches demonstrate the utility of geometric prediction, while aggregate mapping performance alone does not reveal which decisions are caused by unsupported completion.

We focus on this perception--planning interface. Rather than changing the occupancy architecture, we freeze a planner, intervene on its geometry, and trace changes in viewpoint utility and coverage. The proposed online correction layer preserves predictions in sparsely observed space and suppresses those that remain unsupported after repeated nominal exposure. This interface matters when learned completion guides action during long periods of onboard autonomy.

\section{Learned Occupancy in Two Planning Roles}

\subsection{Planner-facing occupancy}
At planning step $t$, the agent has accumulated an RGB-D surface point cloud $\mathcal{S}_t$ and visited poses $\mathcal{C}_t$. A pretrained network predicts an occupancy probability $\hat{\sigma}_t(\mathbf{x})$ for a 3D query point; we threshold it at 0.5 to obtain $\widehat{\mathcal O}_t$. We instantiate the study with MAGICIAN~\cite{li2026magician}. It converts predicted occupied points to Gaussian primitives and renders their novelty from candidate views to estimate coverage gain. The same occupancy is used for collision checking during long-horizon trajectory search. Thus, occupancy changes both a trajectory's score and whether the planner regards it as feasible.

Let $\widetilde{\mathcal O}_t$ denote the occupied set exposed to the planner and $\mathcal T_t$ its candidate trajectories. The shared interface is schematically
\begin{equation}
\boldsymbol{\tau}^{*}_t
=
\arg\max_{\boldsymbol{\tau}\in\mathcal T_t}
\sum_{\mathbf c\in\boldsymbol{\tau}}
G_t(\mathbf c;\widetilde{\mathcal O}_t)
\quad
\mathrm{s.t.}\quad
\operatorname{Free}
(\boldsymbol{\tau};\widetilde{\mathcal O}_t)=1,
\label{eq:planner_interface}
\end{equation}
where $G_t$ is rendered predicted surface gain and $\operatorname{Free}$ is the collision-feasibility test. A single occupied prediction can therefore change the objective, the feasible set, or both. A false positive may look like an uncovered target and may also reject an otherwise traversable route; a false negative may remove a useful target or change inferred connectivity.

These effects unfold in closed loop. The selected trajectory determines the next RGB-D measurements, those measurements update $\mathcal S_t$ and the learned prediction, and the new planner-facing geometry changes the next planning problem. Pointwise occupancy accuracy at one step cannot fully characterize this downstream effect. Our interventions measure the net result after the feedback has unfolded.

\subsection{Controlled interventions}
We hold the occupancy network, candidate views, beam search, collision rule, motion budget, and reconstruction pipeline fixed. Only the occupancy supplied to the planner changes. We compare: (i) accumulated observations only; (ii) learned occupancy with oracle false positives removed; (iii) the learned occupancy baseline; (iv) learned occupancy with oracle false negatives restored; and (v) ground-truth occupancy.

These conditions form a diagnostic decomposition, rather than five separately trained models. With $\mathcal O^{*}$ denoting reference occupied samples and $\mathrm{FP}_t,\mathrm{FN}_t$ the oracle-tagged error sets, the planner inputs are
\begin{align}
\widetilde{\mathcal O}^{\rm obs}_t
 &=\mathcal S_t, \nonumber\\
\widetilde{\mathcal O}^{-\rm FP}_t
 &=\mathcal S_t\cup
   (\widehat{\mathcal O}_t\setminus\mathrm{FP}_t), \nonumber\\
\widetilde{\mathcal O}^{\rm base}_t
 &=\mathcal S_t\cup\widehat{\mathcal O}_t, \nonumber\\
\widetilde{\mathcal O}^{+\rm FN}_t
 &=\mathcal S_t\cup\widehat{\mathcal O}_t\cup\mathrm{FN}_t, \nonumber\\
\widetilde{\mathcal O}^{\rm GT}_t
 &=\mathcal O^{*}.
\label{eq:interventions}
\end{align}
Observation-only planning measures the net value of completion. The two oracle conditions ask whether commission or omission errors dominate while leaving the other error type intact. Ground-truth occupancy removes both and reveals what geometric correctness can achieve when view generation, trajectory search, motion budget, and reconstruction remain fixed.

For oracle tagging, a predicted point is a false positive if it is farther than $\delta=0.03d_{\rm scene}$ from the reference surface; a reference point is a false negative if no prediction lies within the same tolerance. Ground truth is used only for these diagnostic interventions and evaluation, never by the proposed online filter.

Experiments use Macarons++~\cite{li2026magician}: five scenes, five fixed starts per scene, 100 planning actions and 101 camera poses per trajectory. Each condition runs closed loop, so a changed decision produces different future observations and world-model updates. We measure final surface coverage $C_T$, normalized area under the coverage trajectory (AUC), and $S_{70}$, the first step reaching 70\% of the learned baseline's final coverage for the same start. AUC and $S_{70}$ quantify coverage-acquisition efficiency under the fixed motion budget; they do not measure energy directly.

We rerun the complete trajectory for every intervention rather than rescoring a shared log. After the first changed action, conditions receive different observations and no longer share the same map history. Paired differences thus measure the total system-level consequence of changing planner geometry at the same scene and start, including future sensing; they do not attribute every later trajectory difference to one isolated point.

\begin{table}[t]
\centering
\caption{Controlled occupancy interventions. The learned row reports absolute values; other rows are mean paired changes from that baseline. Lower $\Delta S_{70}$ is faster. Oracle rows use ground truth only for diagnosis.}
\label{tab:interventions}
\setlength{\tabcolsep}{1.8pt}
\scriptsize
\begin{tabular}{@{}llccc@{}}
\toprule
Planner occupancy & Group & $\Delta C_T$ & $\Delta$AUC & $\Delta S_{70}$\\
\midrule
Obs. only & All ($n=25$) & $-0.007$ & $-0.038$ & $+4.6$\\
 & Low ($n=5$)  & $+0.100$ & $+0.055$ & $-8.6$\\
 & High ($n=16$)& $-0.050$ & $-0.073$ & $+9.4$\\
\addlinespace[1pt]
Learned $-$ FP$^{\dagger}$ & All & $-0.004$ & $-0.006$ & $+2.5$\\
 & Low  & $+0.020$ & $+0.019$ & $-0.6$\\
 & High & $-0.029$ & $-0.019$ & $+4.2$\\
\addlinespace[1pt]
Learned (abs.) & All & $0.846$ & $0.644$ & $31.6$\\
 & Low  & $0.676$ & $0.485$ & $39.6$\\
 & High & $0.908$ & $0.706$ & $27.4$\\
\addlinespace[1pt]
Learned $+$ FN$^{\dagger}$ & All & $-0.026$ & $+0.031$ & $-3.0$\\
 & Low  & $-0.044$ & $+0.035$ & $-4.0$\\
 & High & $-0.023$ & $+0.030$ & $-1.9$\\
\addlinespace[1pt]
Ground truth$^{\dagger}$ & All & $+0.031$ & $+0.086$ & $-12.7$\\
 & Low  & $+0.095$ & $+0.178$ & $-28.4$\\
 & High & $+0.004$ & $+0.055$ & $-7.4$\\
\bottomrule
\multicolumn{5}{l}{\scriptsize $^{\dagger}$Ground-truth oracle. Low: baseline $C_T<0.75$; high: $C_T>0.85$.}
\end{tabular}
\end{table}

\subsection{Findings}
First, offline ground-truth tags reveal that phantom occupancy contributes 53.5\% and 35.0\% of the rendered gain in two viewpoints selected by the learned planner. The error therefore enters the decision signal, rather than remaining only a map-quality defect.

Second, completion has start-dependent value. Removing it changes pooled final coverage by only $-0.007$, but this average combines opposite behaviors. In the five starts where the baseline ends below 0.75 coverage, observation-only planning changes $C_T$/AUC/$S_{70}$ by $+0.100/+0.055/-8.6$; in the sixteen starts above 0.85, the changes are $-0.050/-0.073/+9.4$. These post-hoc strata describe heterogeneity and do not imply that failure is predictable from the initial pose.

Third, more accurate occupancy is not monotonically better for closed-loop mapping (Table~\ref{tab:interventions}). Oracle false-positive removal fails to improve pooled performance. Restoring false negatives accelerates coverage ($\Delta$AUC $=+0.031$, $\Delta S_{70}=-3.0$) while reducing final coverage by 0.026. Even complete ground-truth occupancy primarily improves efficiency: $\Delta$AUC is $+0.086$ and $S_{70}$ is 12.7 steps earlier, whereas final coverage rises only 0.031. Ground truth is consequently a diagnostic reference for this fixed planner, not an upper bound on endpoint coverage.

This non-monotonicity is consistent with sequential exploration. A local correction can reorder candidate trajectories, after which the robot observes a different subset of the scene. It may accelerate early coverage without improving the final reachable surface, or remove a misleading target that had incidentally carried the robot toward useful nearby geometry. The gap between ground-truth improvements in AUC and final coverage further indicates that candidate-view placement, reachability, and trajectory search remain limiting even after geometry is corrected. World models therefore require decision-level evaluation alongside geometric accuracy.

\section{Observation-Gated Filtering}

Global removal of completion would discard useful guidance in genuinely unobserved space. We instead classify a predicted point as \emph{unsupported} only after repeated frustum inclusion without nearby accumulated surface support. Let $n_t(\mathbf{x})$ count the acquired camera frustums containing prediction $\mathbf{x}$. The filtered set is
\begin{equation}
\mathcal U_t=\left\{\mathbf{x}\in\widehat{\mathcal O}_t\ \middle|\
n_t(\mathbf{x})\geq K,\ 
\min_{\mathbf{p}\in\mathcal S_t}\|\mathbf{x}-\mathbf{p}\|_2>\epsilon
\right\},
\label{eq:filter}
\end{equation}
with $K=2$ and $\epsilon=0.03d_{\rm scene}$. The planner receives $\mathcal S_t\cup(\widehat{\mathcal O}_t\setminus\mathcal U_t)$. Unsupported points contribute neither rendered gain nor collision constraints. The set is recomputed after each observation, so later nearby surface support restores a prediction. The filter requires no ground truth, retraining, or extra rendering pass. Frustum inclusion is a simple exposure proxy; a filtered point is unsupported, not proven false.

Predictions with $n_t(\mathbf{x})<K$ are deliberately retained, even when they lack nearby support, because the robot may not yet have tested them. Repeated nominal exposure converts the continuing absence of support into a correction signal. Recomputing $\mathcal U_t$ rather than permanently deleting points makes the update reversible: later surface evidence can return a prediction to both planner channels. The filter thus preserves completion in unexplored space while allowing accumulated observations to revise the planner-facing map.

\begin{table}[t]
\centering
\caption{Targeted final coverage (mean$\pm$std, three repetitions).}
\label{tab:filter}
\setlength{\tabcolsep}{5.0pt}
\scriptsize
\begin{tabular}{@{}lccc@{}}
\toprule
Start & Learned baseline & Obs.-gated filter & $\Delta C_T$\\
\midrule
Pantheon/2 & $0.309\pm0.021$ & $\mathbf{0.472\pm0.078}$ & $\mathbf{+0.163}$\\
Sestino/2  & $0.836\pm0.108$ & $\mathbf{0.958\pm0.021}$ & $\mathbf{+0.122}$\\
\bottomrule
\end{tabular}
\end{table}

On the two targeted failure-prone starts selected by the diagnosis, observation-gated filtering raises mean final coverage by 0.163 and 0.122 (Table~\ref{tab:filter}). This is a targeted proof of concept; the 25-start study supports the diagnosis, while broader filter evaluation remains necessary.

\section{Implications for Active Perception in Space Robotics}

The closest deployment setting is rover-based surface or subsurface reconnaissance, where limited viewpoints, delayed supervision, and drift in onboard localization make planning depend on spatial predictions that remain revisable. The filter preserves learned structure in untested terrain, then revises it when repeated observations fail to provide nearby surface support. The same interface may also arise in inspection of habitats or orbital structures.

\begin{table}[t]
\centering
\caption{Translation from the controlled study to space-oriented validation. The final column lists required tests, not demonstrated capabilities.}
\label{tab:scope}
\setlength{\tabcolsep}{2.2pt}
\renewcommand{\arraystretch}{1.08}
\scriptsize
\begin{tabular}{@{}p{0.18\columnwidth}p{0.27\columnwidth}p{0.43\columnwidth}@{}}
\toprule
Interface & Current evidence & Required validation layer\\
\midrule
Sensing & Benchmark RGB-D & Low light, dust, noise, and structured dropout\\
State estimation & Benchmark poses & Drift, covariance, and relocalization failure\\
Exposure & Frustum count plus distance & Occlusion- and ray-consistent visibility\\
Mobility & Free 6-DoF motion & Rover kinematics, traversability, and clearance\\
Planner use & One shared filter & Channel-specific safety thresholds\\
Resources & No extra rendering pass & Runtime, memory, energy, and fault response\\
\bottomrule
\end{tabular}
\end{table}

The filter sits after each world-model update and before the planner queries gain or feasibility. Because it uses accumulated observations rather than ground-truth labels or retraining, this interface could operate onboard between communication windows. That placement is attractive for delayed supervision, but the present experiments do not measure latency, memory, energy, or fault recovery. Table~\ref{tab:scope} therefore separates the mechanism tested here from the validation needed for a mission-oriented implementation. The role of onboard autonomy in current Mars rover operations further raises the standard for reliability and operational evidence~\cite{verma2023perseverance}.

Several gaps remain before such a filter can enter a flight-like navigation stack. Under degraded sensing, nominal frustum inclusion is weak evidence. A space-oriented implementation should count exposure only when a prediction is within usable sensor range, is not occluded by supported foreground structure, and receives a valid measurement capable of testing it. A dropped or saturated range return should provide no negative evidence. A valid ray that passes through the predicted location and terminates farther away can instead supply explicit free-space evidence. This distinction matters under low illumination, dust, reflective materials, or range-dependent dropout.

When reliable external localization is unavailable, both frustum membership and nearest-surface distance depend on drifting onboard pose estimates. Pose error can displace a correct prediction from its corresponding measurement, causing the filter to treat it as unsupported. A conservative extension should propagate pose uncertainty into both exposure and support tests, count an exposure only when the point is likely to lie in a valid visible region, and search for supporting observations within an uncertainty-expanded neighborhood. These are deployment requirements; the present study uses accurate benchmark poses and does not evaluate localization drift or failure.

The two planner channels also warrant different evidentiary thresholds. Removing a point from gain estimation changes exploration preference. Removing it from collision checking can admit a new trajectory and carries a greater safety burden. A flight-like system could down-weight gain after repeated valid contradiction while retaining a collision constraint until ray-consistent free space or redundant multi-view confirmation is available. The current shared filter operates on the same coupled planner interface examined in our diagnosis.

Validation should first inject structured depth dropout, range noise, and drift in onboard pose estimates while preserving the same scenes, starts, and planner. Comparisons should separate the present frustum-only gate from depth-aware variants and variants that incorporate pose uncertainty, and report incorrect suppression, collision events, path clearance, latency, and memory in addition to coverage. Planetary-analog trials should then use stereo or depth sensing with onboard visual--inertial or SLAM pose estimates. The present evidence establishes a planner-facing failure and a targeted correction; it does not establish flight readiness.

\section{Conclusion}
Our results motivate rethinking learned occupancy in autonomous active mapping as a revisable planning prior. Because learned occupancy enters both gain estimation and collision checking, its errors can alter where an active mapper looks and which motions it considers feasible. Controlled interventions show that occupancy accuracy alone does not predict closed-loop coverage. Observation-gated filtering retains useful completion while suppressing repeatedly unsupported predictions, improving two targeted failure cases without retraining or ground truth.

Although evaluated here for 3D active mapping, keeping planner-facing predictions revisable also matters in active search. Search-TTA ~\cite{tan2025searchtta} updates its image encoder during search with uncertainty-weighted online gradients, refining potentially inaccurate target-presence maps. Its parameter adaptation differs from our deterministic, reversible 3D occupancy gate, yet both use accumulated task evidence to revise predictions before subsequent planning decisions. We consider active search as a future work.

\sloppy
\hfuzz=1pt
\bibliographystyle{ieeetr}
\bibliography{main}

\end{document}